\documentclass[letterpaper, 10 pt, conference]{ieeeconf}

\IEEEoverridecommandlockouts
\pdfoutput=1
\UseRawInputEncoding

\usepackage{lipsum}
\usepackage{hyperref}
\usepackage{graphicx}
\usepackage{dblfloatfix}
\usepackage{mathtools}
\usepackage{multirow}
\usepackage{booktabs}
\usepackage{array}
\usepackage{float}

\usepackage{amssymb}
\usepackage{pifont}
\newcommand{\cmark}{\ding{51}}%
\newcommand{\xmark}{\ding{55}}%

\title{\LARGE \bf
Lossless SIMD Compression of LiDAR Range \\
and Attribute Scan Sequences
}

\author{Jeff Ford$^{1}$ and Jordan Ford$^{2}$
\thanks{$^{1}$Jeff Ford, ComplexIQ, Atlanta, GA 30517, USA
        {\tt\small jeff@complexiq.com}}
\thanks{$^{2}$Jordan Ford, Department of Electrical Engineering, Carnegie Mellon University,
        Pittsburgh, PA 15217, USA
        {\tt\small jsford@andrew.cmu.edu}}
}

\hypersetup{
    colorlinks,
    allcolors=black
}
\begin{document}

\maketitle
\thispagestyle{empty}
\pagestyle{empty}

\begin{abstract}

As LiDAR sensors have become ubiquitous, the need for an efficient LiDAR data compression algorithm has increased.
Modern LiDARs produce gigabytes of scan data per hour (Fig. \ref{range_bar}) and are often used in applications with limited compute, bandwidth, and storage resources.

We present a fast, lossless compression algorithm for LiDAR range and attribute scan sequences including multiple-return range, signal, reflectivity, and ambient infrared.
Our algorithm---dubbed ``Jiffy''---achieves substantial compression by exploiting spatiotemporal redundancy and sparsity. Speed is accomplished by maximizing use of single-instruction-multiple-data (SIMD) instructions. In autonomous driving, infrastructure monitoring, drone inspection, and handheld mapping benchmarks, the Jiffy algorithm consistently outcompresses competing lossless codecs while operating at speeds in excess of 65M points/sec on a single core. In a typical autonomous vehicle use case, single-threaded Jiffy achieves 6x compression of centimeter-precision range scans at 500+ scans per second.
To ensure reproducibility and enable adoption, the software is freely available as an open source library\footnote[3]{Software is available here: \href{http://github.com/jsford64/jiffy-compression}{http://github.com/jsford64/jiffy-compression}}.

\end{abstract}


\vspace{3mm}

\section{INTRODUCTION}
\begin{figure}[t]
   \centering
   \includegraphics[width=\columnwidth]{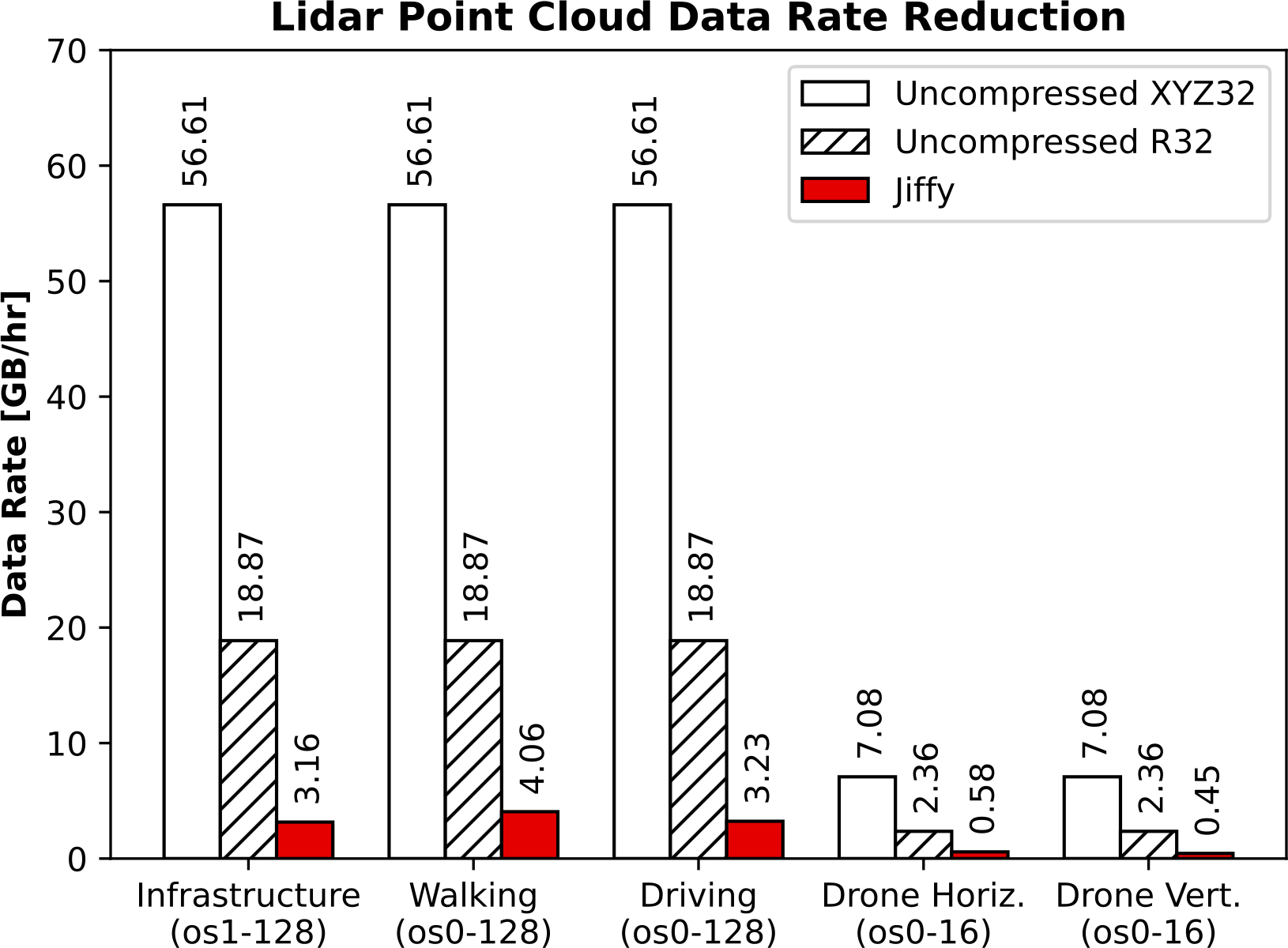}
   \vspace{-3mm}
   \caption{The Jiffy algorithm uses techniques adapted from image encoding and database compression to achieve lossless compression of LiDAR range scan sequences by a factor of 12-18x vs. uncompressed 32-bit Cartesian point cloud format, or 4-6x vs. uncompressed 32-bit range image format in applications including autonomous driving, infrastructure monitoring, and handheld \& drone mapping.
   }
   \label{range_bar}
 \end{figure}

LiDAR sensors are rapidly becoming commonplace in applications ranging from infrastructure monitoring to autonomous cars and drones. At the same time, LiDAR sensors are rapidly improving in range, precision, spatial resolution, and scan rate. Additional outputs, including multi-return range, intensity, reflectivity, and ambient light, result in a flood of LiDAR data that must be processed, transmitted, and stored, consuming scarce resources of bandwidth, memory, time, and power. Fast, lossless LiDAR compression is needed to address the deluge (Fig. \ref{range_bar}).

General purpose algorithms have been used to compress heterogeneous sensor data losslessly, but without prior knowledge of the data, they are limited. Algorithms specifically designed for LiDAR scan data have been proposed to exploit the structure of LiDAR data.
These algorithms have primarily focused on lossy compression of LiDAR range scans by reusing concepts from image, video, or point cloud compression. Thus far, they have proven too slow, too lossy, or too ineffective for wide adoption.

\begin{figure*}[ht]
  \centering
  \includegraphics[height=3.6in]{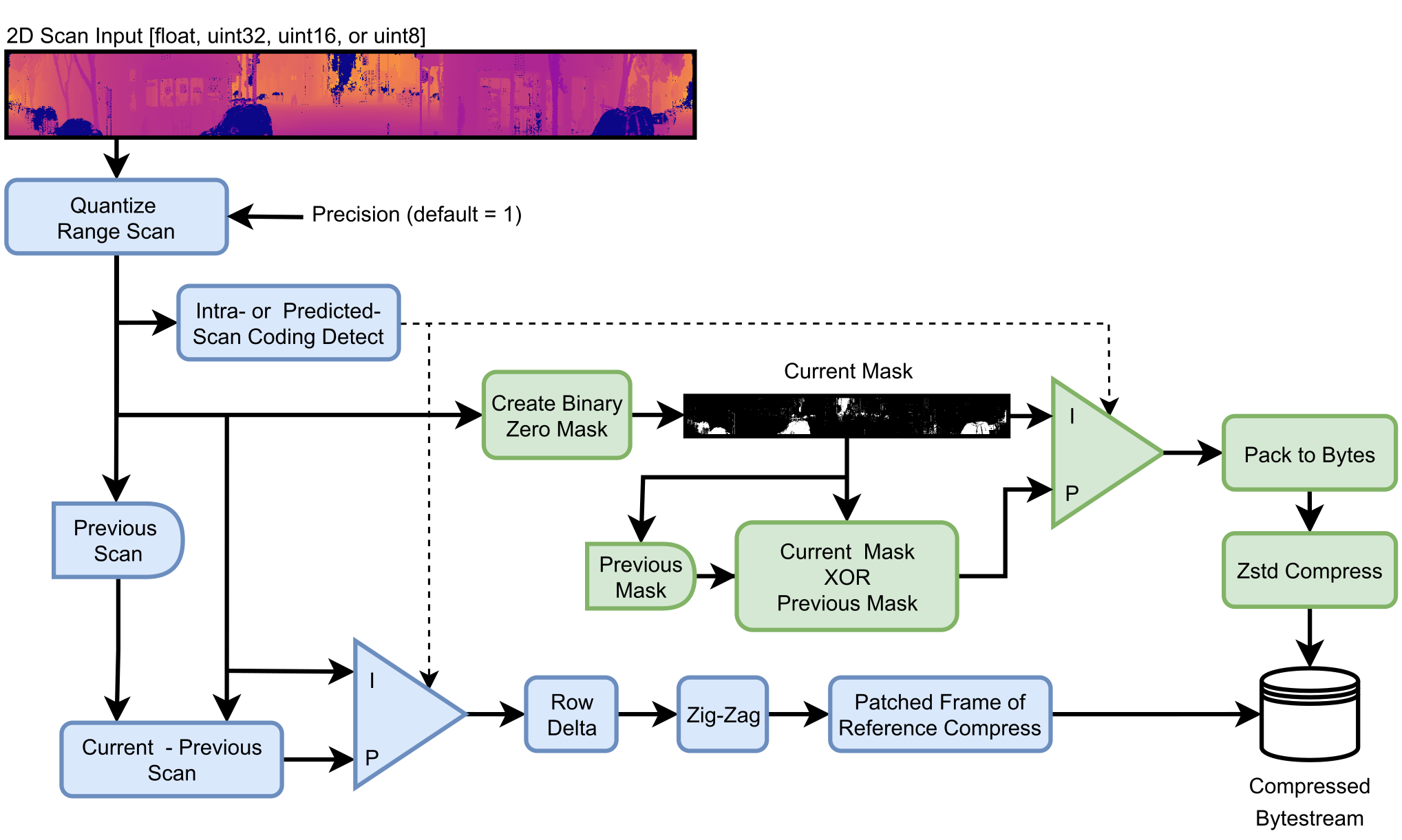}
  \caption{The Jiffy algorithm losslessly compresses high bit-depth, multi-attribute LiDAR scan data. Incoming scans are quantized to the precision of the sensor. A bitmask locating zero-valued samples is extracted from the scan and compressed separately (green) from the remaining scan data (blue). A trial compression heuristic selects an encoding method: intra-scan (I-scan) coding is selected when the previous scan differs significantly from the current scan, otherwise, predicted-scan (P-scan) coding is selected.}
  \label{fast_adaptive_diagram}
\end{figure*}

This work leverages the sparsity and spatiotemporal coherence of LiDAR scan sequences to develop Jiffy--- a fast, lossless codec for high-resolution, multiple-attribute LiDAR scans (Fig. \ref{fast_adaptive_diagram}). Single-threaded performance is ensured by maximizing use of single-instruction-multiple-data (SIMD) instructions. In multi-attribute benchmarks from real-world use cases, Jiffy consistently achieves best-in-class lossless compression with single-CPU throughput in excess of 65 million points per second.

\vspace{4mm}

\section{RELATED WORK}

Some of the earliest general purpose, lossless compression algorithms, Bzip2\cite{SewardBZ2} and DEFLATE (Zlib)\cite{AdlerZLIB} have been used to compress binary files for decades. DEFLATE forms the basis for many application-specific file formats (e.g. ZIP\cite{ZIP_DEFLATE}, DOCX\cite{DOCX}, PNG\cite{PNGSPEC}, OpenEXR\cite{EXRSPEC}).

These early compression algorithms were developed before memory became a bottleneck, before many-core computers became common, and before standard SIMD instruction sets were widely available.
Recently, new algorithms have been designed to take advantage of the hardware in modern CPUs.
LZ4 is a fast dictionary matching algorithm that runs at GB/sec throughput, but produces modest compression ratios\cite{ColletLZ4}.
ZStandard (Zstd) leverages recent developments in Asymmetric Numeral Systems (ANS) and SIMD implementation to achieve improved compression at much faster speeds\cite{ZSTD}.
The popular Robotics Operating System (ROS) initially adopted Bzip2 as its data logging compression codec, and later added LZ4 as a faster option\cite{RosbagCompression}. Both LZ4 and Bzip2 have been replaced by Zstd in ROS2\cite{Rosbag2Compression}. These general purpose compressors are convenient for compressing heterogeneous sensor data because they do not require a specific model of the data, but without a model of the data, their compression ratios are often limited.

Point cloud compression algorithms like Draco\cite{DRACO}, LASZip\cite{LASZIP}, and PCD\cite{DeepPointCloudCompression} have been employed to compress LiDAR-derived point clouds and attributes.
Google's Draco triangle mesh codec treats point clouds as degenerate meshes with no edges or faces. Floating-point vertices are quantized as integers, then ANS coding is applied, similar to Zstd.
The PCD file format groups coordinates by axis, then applies fast LZF compression.
The Motion Picture Experts Group (MPEG) has developed two codecs, G-PCC and V-PCC, for compressing point clouds\cite{mpeg}. The Geometry-Point Cloud Compressor (G-PCC) is based on voxel- and tree-based decomposition of 3D space and is best-suited for static point clouds.
The Video-Point Cloud Compressor (V-PCC) tries to reuse the HEVC video codec. V-PCC searches for planar regions of a point cloud and projects geometry and attribute information onto them. It then packs planar segments into video frames and applies HEVC.
Deep point cloud compression using hardware-intensive neural networks is an active area of research.
A more complete review of point cloud compression can be found in Pereira et al.\cite{PointCloudCodingSurvey}.

The range image representation is a method of representing LiDAR point measurements in spherical coordinates. Conversion from Cartesian XYZ coordinates to the range image representation can mean 3x compression of a LiDAR scan. Azimuth and altitude angles are implied by the x,y position of a sample, leaving an x,y array of polar magnitudes (ranges). Conversion to range images enables the use of image and video compression codecs.
Tu et al. applied JPEG2000 to LiDAR range images\cite{Tu2016Compressing}. Ahn et al. developed a lossless codec for stationary survey scanners\cite{Ahn2015Large-Scale}. Their method recursively subdivides the image into patches and applies a variety of predictors to each patch to achieve reasonable compression at the expense of speed.
Nenci et al. used many instances of H.264 video codecs to compress LiDAR sequences\cite{UseH264}. They achieved 10x lossless compression of Microsoft Kinect range images at 25 Hz using 8 parallel video streams, but their method does not scale to long range sensors. Their method requires 391 concurrent video streams to encode a 100 meter range LiDAR scan with 1 mm resolution.




Researchers interested in database compression have developed integer compression algorithms capable of losslessly compressing billions of unsigned 32-bit integers per second at compression ratios competitive with general purpose compressors like Zstd. Goldstein et al.\cite{FOR} developed Frame Of Reference (FOR) compression, which groups values into frames (e.g. 128 integers). The minimum value in the frame is identified, then the values in that frame are re-coded as an offset from that minimum. Zukowski et al.\cite{PFOR} proposed Patched-FOR (PFOR), which stores values exceeding a thresholded number of bits in a separate list. Lemire et al. \cite{SIMD-FastPFOR} implemented SIMD-FastPFOR (SIMDPFOR)--- PFOR using SIMD acceleration.

\section{METHODOLOGY}

\subsection{Standardizing LiDAR Input}
LiDAR scanners produce geometry and attribute information in formats that differ by manufacturer. Some sensors output range data as an ordered stream of floating point Cartesian coordinates. Others represent ranges as an integer number of millimeters and natively output range images. Signal, reflectivity and other attributes may be returned as floats, as n-byte unsigned integers, or not at all.

Jiffy expects all scan types in a canonical 2D format. Cartesian points must be converted to spherical coordinates, sorted by altitude and azimuth, then quantized to 1, 2 or 4 byte unsigned integers.
Invalid or out-of-range measurements are set to a sentinel value of zero.
The user-provided quantization precision is stored in the compressed bitstream and used to restore the original representation of a quantized scan. Attribute values are similarly arranged and quantized.

\subsection{Quantization}
Quantization precision is an important parameter of the Jiffy encoder. Range measurements are quantized losslessly by default, but lossy quantization can be used to improve compression.
Pereira\cite{PointCloudCodingSurvey} defines quantization as lossless when the quantization precision is less than or equal to the resolution of the sensor. By default, the Jiffy encoder quantizes range measurements using the sensor resolution---one millimeter for all sensors evaluated here (Table \ref{tab:sensor-info}).

In some applications the sensor resolution is orders of magnitude more precise than is necessary (e.g. a streaming visualizer), and aggressive quantization can be used to substantially improve compression. It is straightforward to set the quantization precision, and the maximum quantization error is always less than half of the precision.

\subsection{Intra-Scan (I-Scan) Compression}
Delta compression exploits spatial redundancy by subtracting neighboring samples within a scanline. Jiffy subtracts each sample in a scanline from its left-side neighbor and compresses the results.

Delta encoding is effective on range data because the difference between adjacent samples is typically much smaller than the samples themselves. However, because Jiffy encodes 'out of range' (too far or too close) with a zero-valued sentinel, transitions into- and out-of-range can produce very large deltas, reducing compression.
To eliminate the large deltas caused by these transitions, Jiffy creates a binary mask of one bit per sample. The mask records a zero bit for in-range samples and a one for out-of-range samples. The bitmask is packed into bytes and compressed using Zstd.
After creating the out-of-range bitmask, all out-of-range samples are removed from the 2D scan, creating a flattened 1D vector. The 1D vector can now be delta-encoded without introducing large differences caused by transitions into and out of range.
Masking of zero values is not required for attribute scan types since they do not use zero to indicate valid or invalid measurements. However, Jiffy applies zero masking to all scan types, producing a positive effect on compression ratio.

The SIMD-PFOR algorithm is designed to compress unsigned integers, but delta encoding produces many negative values.
Signed integers can be cast to unsigned and sent directly to SIMD-PFOR for compression, but negative integers cast to extremely large unsigned values because two's-complement sign extension sets all of their highest bits. Since SIMD-PFOR is designed to pack small unsigned integers with many leading zeros in the upper bits, this leads to poor compression. Jiffy uses ZigZag encoding to represent negative differences without the problems caused by sign extension (Equation \ref{eqn:ZigZag}).

\begin{equation}
    \text{ZigZag}(x) = 
    2\lvert x\rvert + \begin{cases*}
                    1 & if  $x < 0$  \\
                    0  &otherwise
                 \end{cases*}
\label{eqn:ZigZag}
\end{equation}

Jiffy performs standalone (Intra-scan or I-scan) compression of individual LiDAR scans by masking and removing out-of-range samples, delta encoding the remaining values, ZigZag encoding the deltas, and SIMD-PFOR compressing the result. I-scan decoding is performed by a straightforward reversal of the encoding procedure.

\subsection{Predicted-Scan (P-Scan) Compression}
When LiDAR sensors are stationary or moving slowly, successive scans exhibit a high degree of similarity that can be exploited to improve compression. Stationary infrastructure monitoring LiDARs typically overlook a largely static scene. Autonomous vehicles are frequently stationary, waiting for passengers or idling in traffic. In these situations, it is redundant to encode near-identical scans of a static scene. Jiffy provides an inter-scan compression mode called Predicted-scan (P-scan). The 'predicted' nomenclature is borrowed from MPEG video compression naming conventions.
Like Intra-scan encoding, a compressed bitmask is created to represent zero-valued samples in the input scan. In predicted-scan (P-scan) mode, the bitmask for the current frame is XORed with the bit mask from the previous frame to remove redundancy from successive bitmasks. The XORed mask result is compressed using Zstd and appended to the output bitstream.
Jiffy's P-scan compression mode uses temporal differencing to avoid encoding redundant scans. After masking the current frame and the previous frame with the current bitmask, the scan differences are encoded using the I-scan compression method.

\subsection{Automatic I/P Mode Selection}
For some applications, I-scan only or P-scan only encoding might make sense. For example, a fast-moving drone could use I-scan only encoding for best results, whereas a pole-mounted monitoring sensor would benefit from using P-scan only. However, applications such as a car in stop-start traffic, a racing drone sitting on its launch pad, or a tree branch waving in front of a static monitor sensor should also be considered. These cases, and in fact, most use-cases benefit from the ability to adaptively select a compression mode on a per-scan basis.
 
The Jiffy encoder could employ brute force to select a compression mode. Because I- and P-scan compression modes are extremely fast, it might be reasonable to encode each scan using both methods and then select the better result. Although brute force trial compression ensures the most effective mode is used, it also reduces throughput by 50\%. As explained in subsection \ref{heuristic}, a trial compression of a few scanlines in each scan (excluding mask compression) proves to be successful in selecting the most effective compression mode for the full scan.

\begin{figure*}[ht]
  \centering
  \includegraphics[height=6.in]{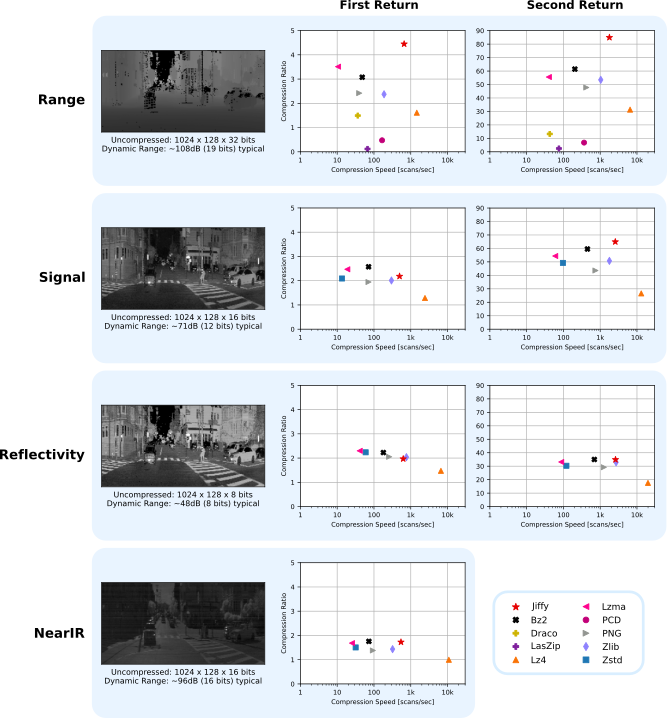}
  \caption{The Jiffy codec compresses LiDAR range scans by a factor of 4.4 at a rate of ~660 scans per second on a single core of a 12th Gen Intel i7-1250U processor. The same codec is pareto-optimal for compression of 16-bit per pixel intensity (Signal) and ambient infrared (NearIR) scan data and is near-optimal for compression of 8-bit per pixel reflectivity data. For sparse second-return data, the Jiffy codec operates at 1000 scans per second and compresses at 35-85x compression ratios. Results presented here are averaged across three autonomous vehicle datasets collected using Ouster OS0-128, OS1-128, and OS2-128 LiDAR sensors.}
  \label{driving_scatter}
\end{figure*}

\section{EXPERIMENTS}

\subsection{Software and System}
All code, including benchmarks, compatible datasets, implementations of competing codecs, and unit tests, are freely available\footnotemark[3].
All codecs and benchmarks are implemented in python3. Performance critical code relies on python3 bindings to fast C/C++ libraries such as numpy, Zstd, SIMD-PFOR, zlib, lz4, etc.
Benchmarks were performed on a 12th Gen Intel i7-1250U CPU and a NVidia Jetson Xavier NX ARM CPU (Table \ref{tab:codec-speed}), both running at 1.9GHz. The benchmark process is pinned to CPU 0 (a performance core) and configured to use the performance CPU governor. Sensors used are summarized in Table 
\ref{tab:sensor-info}.

\subsection{Datasets}
Benchmarks for several LiDAR use cases are employed: autonomous driving, infrastructure monitoring, handheld mapping, and drone flying. To simplify evaluation, all datasets were constructed using Ouster OS0, OS1, or OS2 LiDAR sensors. The autonomous driving and infra- structure monitoring datasets were supplied by the sensor manufacturer\cite{OusterData}. The handheld mapping dataset is extracted from the park sequence of the multi-camera Newer College dataset\cite{zhang2021multicamera}, and the drone datasets are derived from the nya\_02 sequence of the NTU VIRAL dataset\cite{nguyen2021ntuviral}. The NTU drone dataset includes simultaneous scans from one vertically-mounted and one horizontally-mounted OS0-16 LiDAR sensor. Because their compression characteristics differ significantly, the horizontal- and vertical-mounted drone LiDAR scans are treated as separate use cases.

The autonomous driving datasets were collected using the short-range, medium-range, and long-range sensors, detailed in Table \ref{tab:sensor-info}. Results for the codec comparisons in subsection \ref{codec-compare} are averaged across all three sequences.
All datasets contain range scans with single-precision floating point samples. The autonomous driving datasets additionally contain signal, reflectivity, near-infrared, and second-return scan types. The second-return range2 scan type contains 32-bit ranges. The signal scan type and its second-return equivalent, signal2, record the 16-bit unsigned integer intensity of the LiDAR return pulse. The reflectivity and reflectivity2 scan types record an 8-bit unsigned integer estimate of surface reflectivity for every sample in a scan. The near-infrared scan type is a 16-bit unsigned integer measurement of ambient light. Dataset features are summarized in Table \ref{tab:dataset-info}.

\begin{table}[t]
\centering
\caption{A Summary of Benchmarked LiDAR Sensors}
\label{tab:sensor-info}
\resizebox{\columnwidth}{!}{%
\begin{tabular}{@{}lcccc@{}}
\toprule
Sensor & Range & Resolution & Accuracy & Precision \\ \midrule
Ouster OS0   & 0.3 -  \phantom{0}50 m       & 1 mm       & $\pm$ 3 cm & $\pm$ \phantom{0.}1-5 cm  \\
Ouster OS1   & 0.3 - 120 m                  & 1 mm       & $\pm$ 3 cm & $\pm$ 0.7-5 cm    \\
Ouster OS2   & 1.0 - 240 m                  & 1 mm       & $\pm$ 3 cm & $\pm$ 2.5-8 cm    \\ \bottomrule
\end{tabular}%
}
\end{table}

\begin{table}[t]
\centering
\caption{A Summary of Benchmarked Datasets}
\label{tab:dataset-info}
\resizebox{\columnwidth}{!}{%
\begin{tabular}{@{}lcccc@{}}
\toprule
Dataset Name   & Sensor & Scan Shape                 & \# Scans & Attributes            \\ \midrule
Driving-0      & OS0    & 128 $\times$ 1024          & 1432     & \cmark \\
Driving-1      & OS1    & 128 $\times$ 1024          & \phantom{0}719      & \cmark \\
Driving-2      & OS2    & 128 $\times$ 1024          & \phantom{0}971      & \cmark \\
Infrastructure & OS1    & 128 $\times$ 1024          & \phantom{0}690      & \xmark \\
Handheld       & OS0    & 128 $\times$ 1024          & 1947     & \xmark \\
Drone (Horizontal) & OS0 & \phantom{0}16 $\times$ 1024 & 4187     & \xmark \\
Drone (Vertical)   & OS0 & \phantom{0}16 $\times$ 1024 & 4186     & \xmark \\ \bottomrule
\end{tabular}%
}
\end{table}

\subsection{Compression Results vs. Existing Lossless Algorithms\label{codec-compare}}

A comparison of encoding speed and compression ratio for the Jiffy codec and competing lossless methods was performed using the autonomous vehicle data sets\cite{OusterData}.  The mean speed and mean compression ratio over all range scans are presented in Fig. \ref{driving_scatter}.
Lz4 is consistently fastest, but its compression ratio ranks among the worst.
The Jiffy codec is the clear winner in both speed and compression ratio for the first-return 'range' scan types and the second-return 'range2' and 'signal2' scan types. These scan types exhibit the most sparsity and the highest dynamic range, so they benefit most from the zero bitmasking and SIMD-PFOR algorithms. Horizontal delta encoding provides compression gains due to the low spatial frequency content of this scan type, and adaptive coding of I- and P-scans also contributes to higher compression ratios when the vehicle comes to a stop.

Reflectivity (first- and second- returns) were the most challenging for Jiffy, as these scan types had the smallest dynamic range (8 bits) coupled with with the highest spatial frequency content. The high frequency content counters the benefit of delta encoding, and coupling that deficit with the 8-bit dynamic range provides little opportunity for the SIMD-PFOR codec to effectively pack the data. The Zlib codec was a close competitor for these scan types.
The Jiffy compression ratio for the nearIR 16-bit scan type was very competitive with the Bzip2 and Lzma codecs at a ~7x to ~21x higher compression speed.

\subsection{Compression Results by Use Case}

Compression performance was tested for five unique usage scenarios: autonomous driving, with short- medium- and long-range sensors, infrastructure monitoring with a stationary long range sensor, a short range walking scan using a handheld sensor without motion stabilization, and a drone equipped with two short range sensors, one scanning horizontally and one vertically (Fig. \ref{precision}).

For the autonomous driving scenario at the default 1 mm precision, the short range sensor had the best compression ratio (4x), and the long range sensor had the lowest (2.6x). This result clearly demonstrates the efficient bit packing of the SIMD-PFOR algorithm, because the short range sensor data uses fewer of the original 32 bits per sample than the medium- and long-range sensors.

The drone scenario clearly demonstrates the effect of the zero bitmask compression. The horizontally- and vertically-mounted sensors have the same dynamic range, but the vertically mounted sensor data has a higher degree of sparsity relative to the horizontally mounted sensor, so it achieves a higher compression ratio (3.3x horizontal vs. 4.0 vertical).

The infrastructure scenario clearly demonstrates the coding gain of the temporal (P-Scan) compression. This scenario is using the long-range sensor, yet it exhibits a compression ratio that rivals the best-compressed scenarios using the short range sensor, drone(vertical), walking, and driving.

\subsection{Validation of Spatiotemporal Encoder Selection Heuristic}
\label{heuristic}

Testing the efficacy and error rate of the I/P scan selection heuristic was performed by comparing the \textit{predicted} optimal compression method (I or P), to the \textit{actual} optimum compression method for each of 22,877 range scans in all test datasets. The \textit{actual} optimal method was determined by comparing the relative results of compressing the scans using an I-scan only compressor and a P-scan only compressor. Using just 4 test lines in the heuristic test, the algorithm correctly predicted the optimal compression method for 96\% of the 22,877 scans tested, with 0.5\% sub-optimal I-scans and 3.5\% sub-optimal P-scans.

\subsection{Compression / Decompression Speed}

Jiffy encode/decode speed was tested using three autonomous driving datasets. All three datasets use high data rate sensors that produce 128 $\times$ 1024 scans at 10 Hz. Benchmark results for each scan type are presented in Table \ref{tab:codec-speed}. On a single CPU core, the Jiffy codec sustains a mean encoding(decoding) rate of more than 1300(2100) scans/sec over all scan types. On a single thread, Jiffy can encode(decode) all seven scan types at a rate of 120 Hz, twelve times faster than the sensor's 10 Hz output rate.

\begin{table}[h]
\centering
\caption{Jiffy Compression/Decompression Rates by Scan Type}
\label{tab:codec-speed}
\resizebox{\columnwidth}{!}{
\begin{tabular}{@{}lcc@{}}
\toprule
Scan Type     & Intel Encode / Decode [scans/sec] & ARM Encode / Decode [scans/sec]  \\ \midrule
Range         & \phantom{0}661 / \phantom{0}881 &   \phantom{0}595 / \phantom{0}782                  \\
Range2        & 1782 / 3335 &  1602 / 2749                  \\
Signal        & \phantom{0}496 / \phantom{0}579 &   \phantom{0}572 / \phantom{0}743                  \\
Signal2       & 2574 / 4541 &  1777 / 2953                  \\
Reflectivity  & \phantom{0}624 / \phantom{0}799 &   \phantom{0}548 / \phantom{0}631                  \\
Reflectivity2 & 2608 / 4454 &  1767 / 2972                  \\
Near Infrared & \phantom{0}542 / \phantom{0}579 &   \phantom{0}558 / \phantom{0}613                  \\ \bottomrule
\end{tabular}%
}
\end{table}

\begin{figure}[b]
  \centering
  \includegraphics[width=0.8\columnwidth]{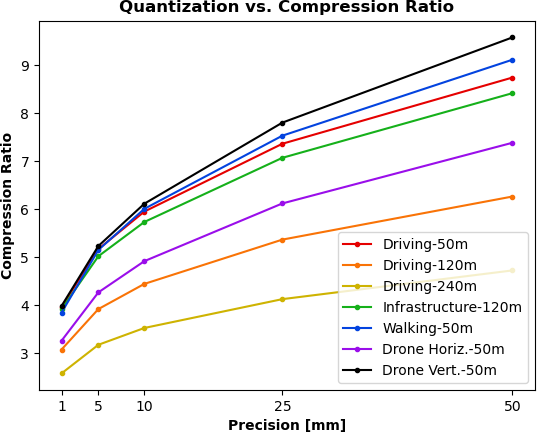}
  \caption{Jiffy quantizes LiDAR range measurements prior to lossless compression. For applications where some error is acceptable, aggressive quantization can be traded for improved compression. Doubling the quantization precision $P_q$ decreases the compressed output by approx. one bit per measurement. Max. roundtrip error is upper bounded by $0.5 P_q$.}
  \label{precision}
\end{figure}

\subsection{Effect of Quantization on Compression Ratio}
Jiffy quantizes LiDAR range data with a precision of one millimeter by default. For less precise sensors, or in circumstances where quantization errors are acceptable, more aggressive quantization can be used to improve compression. Increased quantization leads to a predictable improvement in compression performance with bounded absolute errors (Fig. \ref{precision}).
Doubling the quantization precision removes one bit of information from each range measurement. This enables Jiffy to compress each range measurement using at least one less bit and produces a corresponding improvement in compression ratio. Because Jiffy compression of quantized scan data is lossless, the decoded error will not exceed half of the quantization precision setting.

\subsection{Ablation Study of the Jiffy Algorithm}
An ablation study of the Jiffy encoder using an infrastructure monitoring LiDAR dataset\cite{OusterData} demonstrates the compression achieved by each component of the Jiffy algorithm (Table \ref{tab:ablation-results}). 
The entire dataset is encoded using bare SIMD-PFOR compression, resulting in a 3.2x compression ratio. Adding horizontal delta-encoding along the scanlines and compressing the residuals using SIMD-PFOR predictably lowers the ratio to 2.2x, due to the effect of negative integers on SIMD-PFOR. The ZigZag transform resolves this problem, boosting the ratio to 3.8x.
The out-of-range mask encodes zero values in the range scan using much less than one bit per sample. The bitmask also allows the removal of zero values from the delta-encoded measurement stream. This effectively removes large residuals at the boundaries between out-of-range and in-range values. With masking, delta encoding, the ZigZag transform, and SIMD-PFOR compression, the dataset is compressed by a factor of 5.4x.
Adding the final algorithmic improvement, P-scan prediction, completes the Jiffy encoding algorithm, achieving a final dataset compression ratio of 6.2x.

\begin{table}[h]
\centering
\caption{Performance of Ablated Jiffy Codecs Applied to the Infrastructure Monitoring LiDAR Dataset\cite{OusterData}}
\label{tab:ablation-results}
\resizebox{\columnwidth}{!}{
\begin{tabular}{@{}lc@{}}
\toprule
Ablated Codec Description & Compression Ratio \\ \midrule
SIMD-PFOR     & 3.2               \\
Delta$\rightarrow$SIMD-PFOR  & 2.2               \\
Delta$\rightarrow$ZigZag$\rightarrow$SIMD-PFOR & 3.8               \\
Mask$\rightarrow$Delta$\rightarrow$ZigZag$\rightarrow$SIMD-PFOR   & 5.4               \\
Mask$\rightarrow$Predict$\rightarrow$Delta$\rightarrow$ZigZag$\rightarrow$SIMD-PFOR (Jiffy) & 6.2               \\ \bottomrule
\end{tabular}%
}
\end{table}

\section{CONCLUSIONS}
The Jiffy codec successfully addresses the need for a fast, lossless LiDAR compression algorithm. In multiple-attribute benchmarks from a variety of use cases, Jiffy achieved best-in-class lossless compression at a rate of hundreds of scans per second on a single CPU core. This is more than 10x faster than the uncompressed sensor output rate for all of the sensors tested. The ability to control the quantization of range data is an effective and intuitive way to influence the compression ratio with a predictable error ceiling.

The techniques introduced here could be easily adapted to compress high dynamic range depth images and videos produced by stereo cameras or solid-state LiDAR sensors. Further research could investigate compression of per-pixel velocity measurements produced by continuous-wave LiDAR sensors or of derived perception data such as per-pixel segmentation labels. Lossy compression is also of considerable interest, but research into lossy algorithms should also quantify the downstream impact of compression artifacts on perception and localization algorithms.

\addtolength{\textheight}{-11cm}

\clearpage


\end{document}